\documentclass[conference]{IEEEtran}
\IEEEoverridecommandlockouts
\usepackage{cite}
\usepackage{adjustbox}
\PassOptionsToPackage{bookmarks=false}{hyperref}

\usepackage{amsmath,amssymb,amsfonts}
\usepackage{textcomp}
\def\BibTeX{{\rm B\kern-.05em{\sc i\kern-.025em b}\kern-.08em
    T\kern-.1667em\lower.7ex\hbox{E}\kern-.125emX}}

\usepackage{graphicx}

\usepackage{algorithm}
\usepackage{algpseudocode}

\usepackage{siunitx}
\usepackage{enumitem}
\usepackage{hyperref}

\usepackage{mathtools}
\DeclarePairedDelimiter{\ceil}{\lceil}{\rceil}
\DeclareMathOperator*{\argmin}{argmin} 

\makeatletter
\def\ps@IEEEtitlepagestyle{%
  \def\@oddfoot{\mycopyrightnotice}%
  \def\@evenfoot{}%
}

\def\mycopyrightnotice{%
  {\footnotesize 978-1-7281-0858-2/19/\$31.00~\copyright2019 IEEE \hfill}
  \gdef\mycopyrightnotice{}
}
\@ifundefined{showcaptionsetup}{}{
 \PassOptionsToPackage{caption=false}{subfig}}
\usepackage{subfig}
\makeatother

\usepackage{eso-pic}

\usepackage{footnote}

\newcommand\indentNow{\ \ \ \ }

\begin{document}

\title{FAE: A Fairness-Aware Ensemble Framework}

\author{\IEEEauthorblockN{Vasileios Iosifidis}
\IEEEauthorblockA{\textit{Leibniz University Hanover } \\
\textit{L3S Research Center}\\
Hanover, Germany \\
iosifidis@L3S.de}
\and
\IEEEauthorblockN{Besnik Fetahu}
\IEEEauthorblockA{\textit{Leibniz University Hanover} \\
\textit{L3S Research Center}\\
Hanover, Germany \\
fetahu@L3S.de}
\and
\IEEEauthorblockN{Eirini Ntoutsi}
\IEEEauthorblockA{\textit{Leibniz University Hanover } \\
\textit{L3S Research Center}\\
Hanover, Germany \\
ntoutsi@L3S.de}
}

\maketitle

\begin{abstract}
Automated decision making based on big data and machine learning (ML) algorithms can result in discriminatory decisions against certain protected groups defined upon personal data like gender, race, sexual orientation etc. Such algorithms designed to discover patterns in big data might not only pick up any encoded societal biases in the training data, but even worse, they might reinforce such biases resulting in more severe discrimination.
The majority of thus far proposed fairness-aware machine learning approaches focus solely on the pre-, in- or post-processing steps of the machine learning process, that is, input data, learning algorithms or derived models, respectively. However, the fairness problem cannot be isolated to a single step of the ML process. Rather, discrimination is often a result of complex interactions between big data and algorithms, and therefore, a more holistic approach is required.

The  proposed  \texttt{FAE} (\emph{Fairness-Aware Ensemble}) framework combines fairness-related interventions at both pre- and post-processing steps of the data analysis process. In the pre-processing step, we tackle the problems of under-representation of the protected group (\textit{group imbalance}) and of \emph{class-imbalance} by generating balanced training samples. In the post-processing step, we tackle the problem of \emph{class overlapping} by shifting the decision boundary in the direction of fairness.
\end{abstract}


\section{Introduction}
\label{indroduction}
Machine Learning powered by big data offers incredible opportunities for effective decision making and automation. 
However, several recent incidents have raised concerns about the implications of such systems in terms of fairness~\cite{united2014big}. Amazon's models, to name but one example, that decide which regions of a city are eligible for the prime service, excluded predominantly black ZIP codes in several US cities, like Bronx~\cite{AmazonPrime}. According to Amazon, the \emph{protected attribute} race was not used as a predictor. Nonetheless, there might exist \emph{proxy-attributes} to race which lead to discriminatory decisions. 
Protected attributes and proxies are not the only causes of the problem~\cite{calders2013unbiased}. 
Training data often reflect \emph{societal biases} and are not representative of the population (\emph{sample bias}). Moreover, \emph{system bias} might lead into generation of biased data which result into biased models that further reinforce such discriminatory policies, like in predictive policing~\cite{lum2016predict}.

Despite extensive research work in the area of fairness-aware learning, most of the approaches isolate the problem and its solutions to a single step of the ML process, namely, input data, algorithms or resulting models.
While, we share the view on the importance of working on the main source of bias, i.e., the training data as pointed out by recent work, e.g.,~\cite{zafar2017fairness,calmon2017optimized}, 
we believe that this in itself is insufficient, and that in- and post-processing adjustments are necessary to deal with discrimination.

To this end, we propose the \underline{F}airness-\underline{A}ware \underline{E}nsemble (\texttt{FAE}) framework, a holistic approach that combines pre- and post-processing fairness-enhancing interventions to deal with different bias factors and real-world data complexities, namely group imbalance, class imbalance and class overlap.
At pre-processing, we learn an ensemble of ensembles through a combination of bagging and boosting; the bags are carefully selected via stratified cluster sampling to ensure a balanced group- and class-representation, whereas boosting on each bag forces the classifier to focus on the hard-to-classify examples. 
At post-processing, the decision boundary of the learner is shifted so that the target fairness criterion is fulfilled. Our experiments show that such a joint consideration ensures better fairness- and predictive-performance.


\section{Related Work}
\label{relatedwork}
\textit{Pre-processing} methods aim to tackle discrimination by ``correcting'' the training data to eliminate any biases. Bias can be inherited from the input data, e.g., there might exist proxies to sensitive attributes, or under-represented groups or biased class labels. 
Among the most popular methods in this category are class-label swapping, instance re-weighting, sampling, and instance transformation~\cite{kamiran2012data,iosifidisdealing,calmon2017optimized}. \textit{In-processing} methods modify the learning algorithm to eliminate discriminatory behavior. These interventions are typically learner-specific~\cite{zafar2017fairness,krasanakis2018adaptive,kamiran2010discrimination,dwork2018decoupled,iosifidis2019adafair}.  
For instance, Zafar et al.~\cite{zafar2017fairness} add fairness-related constraints in the objective function of a logistic regression model to account for fairness. 
\textit{Post-processing} methods try to modify the model's predictions or decision boundary in order to ensure fairness~\cite{fish2016confidence,kamiran2010discrimination,pedreschi2009measuring}. Kamiran et al.~\cite{kamiran2010discrimination} propose a fair decision tree learner that combines a fairness-aware splitting criterion with post-processing leaf-relabeling. 
Fish et al.~\cite{fish2016confidence} adjust the decision boundary of a boosting model based on the confidence scores of the misclassified instances. Finally, \textit{class-imbalance} methods aim to deal with skewed class distributions. Over the years, many methods have been proposed such as over-sampling \cite{estabrooks2004multiple}, under-sampling~\cite{drummond2003c4}, synthetic data generation like SMOTE~\cite{he2008adasyn} and boosting~\cite{liu2009exploratory}.

\section{Basic concepts} 
\label{preliminaries}
We consider binary classification with $A=\{A_1,\ldots, A_n\}$ being the attribute space and $Y=\{y^+, y^-\}$ the class attribute. Let $dom(A_i)$ be the domain of $A_i$, and $y^+$ is the \textit{target class}, for example, ``receive a benefit''.
Let $SA \in A$ be a \emph{protected attribute} with $dom(SA) = \{s, \bar{s}\}$; $s$ is the discriminated group (referred to as \textit{protected group}), and $\bar{s}$ is the non-discriminated group (referred to as \textit{non-protected group}).
For instance, $SA=$`\emph{gender}' could be the protected attribute with $s=$`\emph{female}' being the protected group and $\bar{s}=$`\emph{male}' the non-protected.
By combining sensitive attribute $SA$ and class $Y$ values, we define four sub-groups: $s^-$, $s^+$,$\bar{s}^-$,$\bar{s}^+$; e.g., $s^-$ denotes the protected negative group, $\bar{s}^+$ denotes the non-protected positive group etc. 
We assume the following learning challenges: \emph{class imbalance}, that is $|s^+|+|\bar{s}^+| \ll|s^-|+|\bar{s}^-|$; \emph{group imbalance}, that is $|s^+|+|s^-| \ll|\bar{s}+|+|\bar{s}^-|$ as well as class overlap, i.e, the positive class $y^+$  overlaps with the negative class $y^-$.

The goal of a fairness-aware classifier is to learn a function 
$f(\cdot): dom(A_i) \times \cdots \times dom(A_n) \rightarrow Y $, s.t.  $f(\cdot)$ can generalize well to unseen instances and does not discriminate against the protected group for the target class $y^+$.

\textbf{Discrimination measure:}
We adopt the \emph{equal opportunity measure} (EQOP)~\cite{hardt2016equality}  
that compares the probability of being predicted as positive  while belonging to the positive class (TPR) between protected $s$ and non-protected $\bar{s}$ groups:
\begin{equation}
EQOP: P\Big(f(d)=y^+| \bar{s}^+\Big) - P\Big(f(d)=y^+| s^+\Big)
\label{eq:EQOP}
\end{equation}\normalsize
$EQOP \in [-1, 1]$: a value close to 0 means \emph{fair outcomes}, and is desirable, whereas a value close to 1 indicates \textit{discriminatory} behavior towards the protected group. A value close to -1 indicates \textit{reverse discrimination} towards the non-protected group.
A classifier $f(\cdot)$ is said to \emph{not discriminate} if: $\mid EQOP \mid \leq \epsilon$. The user-defined threshold $\epsilon$ controls how much discrepancy between the two groups is tolerated. 

\textbf{Predictive performance measure:}
The vast majority of existing works minimize the standard error rate, e.g.,~\cite{zafar2017fairness,kamiran2012data,calmon2017optimized,kamiran2010discrimination,fish2016confidence}, which is not useful in case of \textit{class-imbalance} as it mainly reflects the performance of the model in the majority class. Moreover, EQOP measure, (c.f., Equation\ref{eq:EQOP}) which relies on the TPR difference, is oblivious to the problem of class imbalance.
As an extreme case, if a classifier totally rejects the minority (positive) class and correctly classifies the majority (negative) class then, based on EQOP, the classifier is both fair (in terms of EQOP) and accurate (in terms of error rate). 
Recent methods fall in this pitfall and their low reported discrimination scores are mainly due to low TPR values (c.f., Section~\ref{sec:results}). Hence, we use balanced accuracy~\cite{brodersen2010balanced}:
\begin{equation}\label{eq:balanced_error}
B.ACC = \frac{1}{2}\cdot(\frac{TP}{TP + FN} + \frac{TN}{TN + FP})
= \frac{(TPR + TNR)}{2}
\end{equation}\normalsize

Our approach resembles the EasyEnsemble approach~\cite{liu2009exploratory}, which we adapt for group as well as class imbalance. Specifically, we combine \emph{bagging} and \emph{boosting}; thus, the final model is an ensemble of ensembles. \textit{Bagging} reduces model variance by generating multiple models from bootstrap samples drawn from the training data. \textit{Boosting} reduces both (model) bias and variance by combining many weak learners, each focusing on missclassified examples from  previous learners~\cite{schapire1999brief}.

\section{FAE - A Fairness-Aware Ensemble Framework}
\label{ouralgorithms}
\begin{figure*}[ht!]
 \centering
 \includegraphics[width=1\textwidth]{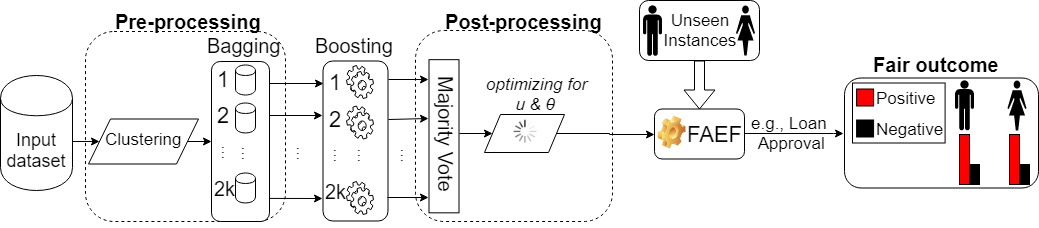}
 \caption{An overview of our holistic pre- and post-processing \texttt{FAE} framework}
 \label{fig:architecture}
\end{figure*}

Figure~\ref{fig:architecture} shows an overview of FAE, from training (left side) to prediction of new instances (right side). FAE combines pre- and post-processing fairness-related interventions, as follows:
\begin{itemize}
    \item \textbf{Fairness-aware ensemble learning} \\
    In \emph{pre-processing}, we tackle the problems of group- and class-imbalance. In particular, we employ \emph{bagging} to \emph{balance} the groups in each bag by taking into account the protected positive group, and a representative sample from the other groups (Section~\ref{sec:fair_bags}). Afterwards, \emph{boosting}~\cite{schapire1999brief} is employed on each bag, so at the end, an ensemble of ensembles is learned. 
\item \textbf{Fairness-aware decision boundary shift} \\ 
In the \emph{post-processing}, we shift the decision boundary of the learner in the direction of fairness based on a tunable parameter $\theta$, until the $EQOP$ score satisfies the user-defined threshold $\epsilon$ (Section~\ref{sec:param_tunning}). 
\item \textbf{Selecting the shortest hypothesis}
Finally, we select the optimal number of boosting models $u \in [k,2k]$ that exhibits the best performance in terms of both fairness and balanced error (Section~\ref{sec:optimization}).
\end{itemize}


\subsection{Fairness-aware ensemble training}
\label{sec:fair_bags}
In the pre-processing step, we tackle discrimination in the training data caused by group and class imbalance ensuring that the protected positive group will also be learned by the model. For that, we propose a fair and representative sample generation process. Each sample is created s.t it contains the whole \emph{protected positive group $s^+$} and a \emph{representative equisized} sample from each of the other groups (i.e., from $s^-, \bar{s}^+, \bar{s}^-$).

Algorithm \ref{alg:training} shows the different steps in the ensemble's training phase. Clustering is applied in the beginning for each group $s^-, \bar{s}^+, \bar{s}^-$ (line 2). We employ \emph{stratified sampling} to ensure a balanced representation, where the \emph{strata} correspond to clusters\footnote{Clustering better approximates the underlying data distributions, accounting for sub-groups, and thus ensuring representative samples from each group.} extracted through some clustering algorithm from the other groups $s^-, \bar{s}^+, \bar{s}^-$. The bags are created (lines 6-7) by combining $s^+$ and a stratified sample from the generated clusters for each group. 
In each bag, an AdaBoost classifier is trained (line 8) and added to the ensemble (line 9).
The output model is an ensemble of ensembles $E$ (line 12):
\begin{equation}
\label{eq:ensemble}
E(x) = \sum_{i=1}^{2k} \bigg(\sum_{j=1}^{z}\Big(a_{i,j}h_{i,j}(x)\Big)\bigg)
\end{equation}\normalsize
where $k$ is the number of bags (c.f., Eq.~\ref{eq:kappa_estimation}), $z$ the number of boosting rounds and $a_{i,j}$ is the weight of the weak learner $h_{i,j}$ ($a$ and $h$ are obtained through AdaBoost).

\subsubsection{\textbf{Stratified sampling}}
\label{sec:bags}
The goal is to generate the different bags $s^{-'}, \bar{s}^{+'}, \bar{s}^{-'}$ from the majority groups $s^-, \bar{s}^+, \bar{s}^-$, respectively, such that: 
$|s^+|=|s^{-'}|= |\bar{s}^{+'}| = |\bar{s}^{-'}|$. 
To ensure representative samples from each group, we cluster each group (i.e., each of $s^-, \bar{s}^+, \bar{s}^-$) and use the resulting clusters for bag generation. 
Note that clusters are generated only once in the beginning of the training process (line 2, Algorithm~\ref{alg:training}) and re-used afterwards.

\subsubsection{\textbf{Estimating the initial number of bags}}
\label{sec:k} The number of bags $k$ must be sufficient to overcome the drawback of potential loss of useful information due to under-sampling (i.e., each bag is a sample of the training data). We overcome this drawback by estimating the number of bags $k$ s.t. we insure that the clustered instances are at least in one of the bags. We calculate the number of bags $k$  as following:
\begin{equation}
\label{eq:kappa_estimation}
k = \ceil[\Bigg]{\frac{max\{|s^-|, |\bar{s}^+|, |\bar{s}^-|\}}{|s^+|}} + 1
\end{equation}\normalsize

In other words, $k$ provides an estimation that an instance from the most populated group will be at least in one bag, thus, avoiding the under-sampling drawback. In practice, we train the ensemble with twice the amount of bags ($2k$ bags); at the post-processing step, we select the best set of learners for the ensemble (Section~\ref{sec:optimization}). 
 
\begin{algorithm}[t!]
 \caption{Pre-processing step}\label{alg:training}
 \begin{flushleft}
 \hspace*{\algorithmicindent} \textbf{Input:} Training set $D$, target class $y^+$, $SA$, $k$\\
 \hspace*{\algorithmicindent} \textbf{Output:} Ensemble E 
 \end{flushleft}
 \begin{algorithmic}[1]
 \State Extract groups $s^+, s^-, \bar{s}^+, \bar{s}^-$ based on $y^+$ and $SA$ from $D$;
 \State Generate clusterings $C_{s^-}, C_{\bar{s}^+}, C_{\bar{s}^-}$ from $s^-, \bar{s}^+, \bar{s}^-$, respectively;
 \State Ensemble $E\gets \{\emptyset\}$; 
 \State $i\gets 1$;
 \For{\texttt{$i=1:2k $}}  
 \State Stratified sample $s^{-'}_i, \bar{s}^{+'}_i, \bar{s}^{-'}_i$ from $C_{s^-}, C_{\bar{s}^+}, C_{\bar{s}^-}$;
 \State Bag $B_i = s^+ \cup s^{-'}_i \cup \bar{s}^{+'}_i \cup	\bar{s}^{-'}_i$;
 \State Train an AdaBoost classifier $H_i$ upon $B_i$;
 \State $E\gets E \cup H_i$;
 \State $i\gets i+1$;
 \EndFor
 \State \textbf{return} ensemble $E$;
 \end{algorithmic}
\end{algorithm}\normalsize

\subsection{Fairness-aware decision boundary tuning}
\label{sec:param_tunning}
Despite the pre-processing interventions, the resulting model $E$ might not fulfill the discrimination threshold $\epsilon$. In FAE, if EQOP $>\epsilon$, a post-processing procedure is invoked that shifts the decision boundary based on a parameter $\theta$ s.t. EQOP $\leq \epsilon$. 

As we show in Section~\ref{sec:results}, by employing only the pre-processing step, the discrimination is significantly reduced. However, a post-processing step is necessary given that discrimination can stem from other factors including class overlap and the accuracy-oriented objective function of Adaboost.

\textbf{Parameter tuning.} For a $SA$ (e.g. $SA=$`\emph{gender}') our goal is to find the optimal threshold parameter $\theta_{s}$ or $\theta_{\bar{s}}$ (for the different attribute values $dom(SA)=\{s, \bar{s}\}$) to minimize $EQOP$. Furthermore, at any given time our ensemble learner $E$ can discriminate against only one of the group $s$ or $\bar{s}$.

Algorithm~\ref{alg:threshold} shows the detailed steps for tuning the optimal $\theta_{s}$ and $\theta_{\bar{s}}$. 
To begin with, we compute the $EQOP$ score, which represents the difference between true positive ratios between $s$ and $\bar{s}$ (line 6). Next, we sort the misclassified instances from $s^+$ and $\bar{s}^+$ groups (lines 7 -- 8) in a descending order (w.r.t the target class) based on their ensemble classification score from Equation~\ref{eq:ensemble}. 
In case $EQOP$ score is below the discrimination threshold $\epsilon$, then $\theta_{\bar{s}} = \theta_{s} = 0.5$ (lines 9 -- 10). Setting the threshold parameter to 0.5 has no implication in classifying test instances in Equation~\ref{eq:ensemble_disc}. For $|EQOP| > \epsilon$, we distinguish between \emph{discrimination} and \emph{reverse discrimination} (lines 11 -- 17). That is, for $EQOP > 0$ the model discriminates against instances with $SA=s$, otherwise against instances with $SA=\bar{s}$. The threshold parameter $\theta_s$ or $\theta_{\bar{s}}$ represents the $E(d)$ score of the last instance from the $top_k$ necessary instances from $MC_{s^+}$ or $MC_{\bar{s}^+}$ (lines 12 and 15) that need to be classified correctly to fulfill the criteria $|EQOP|\leq \epsilon$. 
The $top_k$ instances needed for minimizing the discrimination are obtained as following:
\begin{equation}\footnotesize 
\begin{aligned}
 \frac{top_k + TP_s}{TP_s + FN_s} = \frac{TP_{\bar{s}}}{TP_{\bar{s}} + FN_{\bar{s}}} \Rightarrow top_k = \ceil[\Bigg]{\frac{TP_{\bar{s}}(TP_s + FN_s)}{TP_{\bar{s}} + FN_{\bar{s}}} - TP_s}
 \end{aligned}
\end{equation}\normalsize
where $TP$ and $FN$ stand for true positive and false negative instances of protected and non-protected group respectively.

\begin{algorithm}[t!]
 \caption{Post-processing step}\label{alg:threshold}
 \begin{flushleft}
 \hspace*{\algorithmicindent} \textbf{Input:} $D$, $E$, $s,\bar{s}$, $\epsilon$\\
 \hspace*{\algorithmicindent} \textbf{Output:} $\theta_{s}, \theta_{\bar{s}}$
 \end{flushleft}
 \begin{algorithmic}[1]
 \State $\theta_{s} = \theta_{\bar{s}} = 0.5 $

 \State $MC_{s^+}$, $MC_{\bar{s}^+} \gets \{\emptyset\}$
 \State True positive rate $TPR_s$ and $TPR_{\bar{s}}$ for $s$ and $\bar{s}$
 \State $CC_{s^+} =$ \#correctly classified instances in $s^+$
 \State $CC_{\bar{s}^+} =$ \#correctly classified instances in $\bar{s}^+$
 \State $EQOP = TPR_{\bar{s}} - TPR_s$
 \State Misclassified instances $MC_{s^+}$ and $MC_{\bar{s}^+}$ for $s^+$ and $\bar{s}^+$
 \State Sort $MC_{s^+}$, $MC_{\bar{s}^+}$ in descending order based on $E(d)$ 
 
 \State {\textbf{IF} $|EQOP| \leq \epsilon$} // no discrimination
 \State \indentNow $\theta_s = \theta_{\bar{s}} = 0.5$

 \State {\textbf{ELSE IF} $EQOP > 0$} // discrimination
 \State \indentNow $top_k = \frac{CC_{\bar{s}^+}}{|\bar{s}^+|}|s^+| - CC_{s^+}$
 \State \indentNow $\theta_{\bar{s}} = MC_{s^+}[top_k]$

 \State {\textbf{ELSE IF} $EQOP < 0$} // reverse discrimination
 \State \indentNow $top_k = \frac{CC_{s^+}}{|s^+|}|\bar{s}^+| - CC_{\bar{s}^+}$
 \State \indentNow $\theta_s = MC_{\bar{s}^+}[top_k]$ 
 
 \State \textbf{ENDIF}
 \State \textbf{return} $\theta_{s}, \theta_{\bar{s}}$
 \end{algorithmic}
\end{algorithm}\normalsize

\subsection{Hypothesis selection}
\label{sec:optimization}
Out of the $2k$ learners, we select the shortest hypothesis (in terms of number of bags) that  optimizes the following objective function:

\begin{equation}\label{eq:argmin}
\argmin_u ~ (B.ERR_u + 2\cdot |EQOP_u|)
\end{equation}
where $B.ERR$ is the balanced error rate and $u\in [k,2k]$ is a set of AdaBoost models (each AdaBoost is trained upon a different bag). 
The objective function is applied after the decision boundary adjustment i.e., Algorithm~\ref{alg:threshold} is taking place after the pre-processing step, and afterwards the set of learners that minimize Equation~\ref{eq:argmin} is selected. Since class imbalance is tackled in the pre-processing step, more emphasis is given to the ensemble's fairness in the objective function. The final model (FAE) is: $$E(x) = \sum_{i=1}^{u} \Big(\sum_{j=1}^{z}\Big(a_{i,j}h_{i,j}(x)\Big)\Big)$$.

\subsection{FAE Classification}
\label{sec:classification}
In classifying instances with \emph{FAE}, we distinguish two cases. 
If $|EQOP|\leq \epsilon$, the classification is done solely through the majority voting scheme in $E(d)$ (c.f., Equation~\ref{eq:ensemble}). This is the case, where no post-processing tuning is required, rather pre-processing interventions are adequate in fulfilling the $EQOP$ threshold. For $EQOP<0$ and $|EQOP|>\epsilon$, our model discriminates against $SA=\bar{s}$ in the training set, hence, instances will be classified based on Equation~\ref{eq:ensemble_disc}.
\begin{equation}
\label{eq:ensemble_disc}
f(d) =
 \begin{cases}
 \text{$y^+$} & \quad \text{if d(SA) = $\bar{s}$ and } \text{$E^+(d) \geq \theta_{\bar{s}}$}\\
 \text{E(d)} & \quad \text{otherwise.}
 \end{cases}
\end{equation}\normalsize
where $E^+$ is the probability of $d$  assigned to $y^+$. Similar is the case for $EQOP>0$ and $|EQOP|>\epsilon$; in this case, Equation~\ref{eq:ensemble_disc} is altered by replacing $d(SA)=\bar{s}$ to $d(SA)=s$ and $\theta_{\bar{s}}$ to $\theta_s$.

\section{Experimental Setup}
\label{experiments}
Our framework\footnote{\url{https://iosifidisvasileios.github.io/Fairness-Aware-Ensemble-Framework/}} has been instantiated with Logistic Regression as base learners. Each dataset is randomly split into train $(2/3)$ and test set $(1/3)$ (holdout evaluation, similar to~\cite{zafar2017fairness}). We report on the average of 10 random splits. We set $\epsilon=0$ as a threshold for EQOP (no discrimination). For AdaBoost, the maximum number of boosting rounds $z$ is set to 25. We evaluate the following aspects: (i) classification performance based on 
balanced accuracy (B.ACC, Equation~\ref{eq:balanced_error})
and (ii) discriminative performance based on  \textit{EQOP} (Equation~\ref{eq:EQOP})
. 

\subsection{Datasets}
\label{sec:datasets}
We evaluate our approach with two well known datasets: Adult census income and {Bank. \textbf{Adult census income} dataset~\cite{dua2017} contains demographic data from the U.S. The task is to determine if a person receives more than 50K dollars annually. We use as the target class, people who receive more than 50K per year. We remove duplicate instances and instances containing missing values which results to 45,175 instances. We consider as protected attribute $SA=Gender$ with $s=female$. \textbf{Bank} dataset~\cite{dua2017} is related to direct marketing campaigns of a Portuguese banking institution and contains 40,004 instances. The task is to determine if a person subscribes to the product (bank term deposit). As target class we consider people who subscribed to a term deposit. We consider as $SA=maritial~status$ with $s=married$.

\subsection{Baselines and FAE Ablations}
\label{sec:baselines}


\subsubsection{\textbf{Baselines}}\hfill

\noindent\textbf{Shifted Decision Boundary (\textbf{SDB}) ~\cite{fish2016confidence}:} SDB uses a set of base classifiers in an AdaBoost classifier. Instead of majority voting (i.e., $\sum_{i=1}^{T}a_i h_i(x)$), SDB employs confidence scores (i.e., $\frac{\sum_{i=1}^{T}a_i h_i(x)}{\sum_{i=1}^{T}a_i}$) for predictions. The best threshold value for a specific protected group is established to minimize statistical parity. The shift in the boundary takes place after the training phase, thus, making it a post-processing method and suitable for comparison. To have a fair comparison, we find the best threshold estimation of SDB for EQOP, instead of statistical parity as in the original paper.

 
\noindent\textbf{Disparate Mistreatment (DM):} Zafar et al.~\cite{zafar2017fairness} formulate the fairness problem as a set of constraints, for which they optimize a logistic regression (LR) model. They consider three sets of constrains: (i) minimize difference in FPR (false positive rate), (ii) minimize difference in FNR (false negative rate), and (iii) a combination of both. For our comparison, we employ only (ii) since $TPR = 1 - FNR$. We employ the method's default parameters. 


\noindent\textbf{AdaBoost:} here we consider an ensemble learner (equipped with LR as a weak learner) without any pre- or post-processing fairness-related interventions. The goal is to show the ability of these ensembles to classify under group and class imbalance and its impact on discrimination scores like EQOP.

\noindent\textbf{EasyEnsemble:} EasyEnsemble~\cite{liu2009exploratory} is an ensemble that employs bagging and AdaBoost to tackle class imbalance, with LR as a weak learner. We employ EasyEnsemble to compare our approach with a method that directly tackles class imbalance. We set as number of bags to $N=20$.

\subsubsection{\textbf{FAE Model Ablation}}\hfill

FAE is a joint framework of pre-and post-processing interventions. We consider the following ablations, to evaluate the individual effect of the pre- and post-processing interventions:

\noindent\textbf{Only Bagging (OB)} is the pre-processing step in FAE  (c.f. Section~\ref{sec:fair_bags}). We use OB to show the behavior of the ensemble that is trained upon fair and representative groups, without further tuning its decision boundary. 

\noindent\textbf{Simple Majority Threshold (\textbf{SMT})} refers to the post-processing part in FAE (c.f. Section~\ref{sec:param_tunning}).
This method is similar to SDB~\cite{fish2016confidence}, however, instead of using confidence scores, we use the default majority vote of an AdaBoost classifier. That is, after training, we compute the best parameter $\theta$ for a specific protected group to minimize EQOP (Algorithm~\ref{alg:threshold}). We use SMT to show how individual post-processing tuning affects the performance of the models.

We use EM and K-means clustering algorithms to compare the impact of clustering in the bagging step in FAE and its pre-processing step OB, which we indicate with FAE (EM) and OB (EM), and FAE (K-means) and OB (K-means), respectively. For EM, the optimal number of clusters for each group is estimated via cross validation (100 iterations) while for K-means we use the elbow metric (least squares), where the number of clusters ranges in $[2,25]$.

\section{Evaluation Results and Discussion}
\label{sec:results}
We report on: (i) \emph{classification performance} w.r.t \emph{B.ACC} and (ii) \emph{fairness performance} w.r.t. EQOP. 

\begin{table}
 \centering
  \begin{adjustbox}{width=1\columnwidth,center}
 \begin{tabular}{l l l l l}
 \hline
 & \multicolumn{2}{c}{\emph{Adult Cen.}} 
 & \multicolumn{2}{c}{\emph{Bank}} \\
 \hline
Approach & \textbf{B.ACC. (\%)} & \textbf{EQOP (\%)} & \textbf{B.ACC. (\%)} & \textbf{EQOP (\%)}\\
 \hline
 \textbf{AdaBoost} & 76.56 & 11.92 & 66.32 & -6.25 \\
  \textbf{EasyEnsemble} & 80.58 & 15.72 & 83.24 & -4.52 \\
\textbf{DM} & 70.96 & -11.83 & 65.69 & -0.97  \\
\textbf{SDB} & 77.02 & -2.72 & 66.23 & -5.88  \\
\textbf{SMT} & 76.86 & -2.99 & 73.26  & 30.58  \\
 \textbf{OB (EM)} & 80.91 & -4.31 & 83.10 & 2.21 \\
  \textbf{OB (K-means)} & 80.92 & -4.70 & 83.10 & 1.89\\
 \textbf{FAE (EM)} & \textbf{81.09} & \textbf{1.52} & \textbf{83.29} & \textbf{-0.12} \\
  \textbf{FAE (K-means)} & 81.01 & 1.67 & 83.24 & 0.24 \\
\hline
\end{tabular}
\end{adjustbox}
 \caption{Evaluation results for $B.ACC.$ and $EQOP$. EQOP is in the range of [-1,1], in this case we show the percentage points. The best results are marked in boldface.}
 \label{tab:eval_B.ACC._eqop_results}
\end{table}
 
Table~\ref{tab:eval_B.ACC._eqop_results} shows the scores for the B.ACC metric for both datasets and approaches under comparison. 
Our approach FAE achieves some of the highest $B.ACC$ scores, with an average score of $\overline{B.ACC}=82.19\%$ across all datasets for FAE (EM). Similar is the score of EasyEnsemble with $\overline{B.ACC}=81.91\%$. Yet, in terms of EQOP EasyEnsemble produces highly discriminatory results, since it focuses solely on predictive performance.  

A detailed inspection across the competing approaches reveals that the differences between non-bagging and non-ensemble approaches are highly significant. An even representation of all groups is important for classification performance. For models like AdaBoost, SMT, SDB, DM that do not account for the group imbalance, we see a huge drop in B.ACC scores. FAE (EM) has a 20\% relative increase when compared against DM, and 15\% relative increase against the other models.
 
\noindent{\textbf{Ensemble Learners:}} The case of AdaBoost shows that using solely ensemble learners is not sufficient to ensure a non-discriminatory classification. It has the second lowest performance with $\overline{B.ACC}=71.44\%$. EasyEnsemble which focus on class imbalance has very good predictive performance with $\overline{B.ACC}=81.91\%$; however, this is not sufficient to tackle discrimination. Same behavior can be observed for OB. This confirms our assumption, that such discriminatory behaviors are a result of other factors such as class overlap.

\noindent{\textbf{Bagging:}} Bagging ensures even representations of the different groups, thus, it enables models that achieve better B.ACC. Models that employ bagging achieve similar B.ACC scores. Comparing against other non-bagging approaches, such as AdaBoost, DM and SMT, we note a significant drop in terms of B.ACC. However, it is important to note that a high B.ACC score is not sufficient for non-discriminatory classification behavior because, discrimination is often manifested in terms of uneven probabilities for granting a benefit to different groups (c.f. Section~\ref{preliminaries}). 

Regarding discrimination, we observe that high B.ACC scores do not necessarily correlate with low EQOP scores, that is, discrimination free classification behavior. In our choice of competitors, it is evident that such strategies are often insufficient in minimize discrimination.  

From the competitors, only AdaBoost and EasyEnsemble have low EQOP scores. EasyEnsemble is particularly interesting; its B.ACC score is on average close to FAE (EM), however, it exhibits a high discrimination score with $\overline{EQOP}=10.12\%$. This highlights that optimizing only for classification performance is subject to pitfalls of uneven distributions of groups. Whereas our models, the pre-processing stage OB, and FAE, achieve the lowest discrimination results. FAE (EM) has the lowest score with $\overline{EQOP}=0.82\%$ with nearly an ideal EQOP score.

Contrary, for models that optimize for discrimination free classification, we note a significant decrease of EQOP scores compared to AdaBoost and EasyEnsemble. For example, DM in its optimization function minimizes for the EQOP score, leading to $\overline{EQOP}=8.18\%$. Yet, its B.ACC score is severely impacted. This is mostly due to the fact that it learns a logistic regression model under high group imbalance.

An important comparison is between FAE and DM. FAE provides a high relative decrease of 90\% in terms of EQOP. This shows, that despite the fact that DM optimizes the training objective to reduce discrimination, the impact of fair and balanced representations of all groups in training supervised models is highly important.

\section{Conclusions and Future Work}
\label{coclunsions}
In this paper, we addressed the problem of discrimination against marginal groups in classification models caused by group imbalance, class imbalance and societal encoded biases manifested as class overlap esp. for the protected group. 
We presented the FAE framework, a holistic approach to fairness-aware classification that combines pre-processing balancing strategies with post-processing decision boundary adjustment. The pre-processing stage, which computes the number of bags and determines the different groups and clusters to ensure fair representation allows the models to learn representative classifiers that significantly increase the performance and at the same time reduce the discrimination. Due to the encoded societal biases (\emph{class overlap}) in the data, even representations among groups are insufficient in addressing discrimination. Hence, we shift the decision boundary and additionally select hypotheses from the ensemble learners for nearly ideal  EQOP scores. Such steps ensure that a reduction in terms of EQOP does not come at the cost of the ability of the model to correctly classify instances into their corresponding classes.

Our experiments show that discrimination free models are feasible, and for a given \emph{feature space}, we can achieve maximal classification performance, and account for important factors like discrimination for a given target measure, e.g., EQOP. In our current version of FAE, we employ pre- and post-processing fairness-enhancing interventions. Furthermore, improvements are possible by including in-processing interventions at the algorithm level, thus targeting the whole ML process from data to algorithms and models.

\section*{Acknowledgment}
\addcontentsline{toc}{section}{Acknowledgment}
This work is part of a project that has received funding from the European Union’s Horizon 2020, under the Innovative Training Networks (ITN-ETN) programme Marie Skłodowska-Curie grant (NoBIAS-Artificial Intelligence without Bias) agreement no. 860630. The work is also inspired by the Volkswagen Foundation project BIAS ("Bias and Discrimination in Big Data and Algorithmic Processing. Philosophical Assessments, Legal Dimensions, and Technical Solutions") within the initiative "AI and the Society of the Future"; the last author is a Project Investigator for both of them.

\bibliographystyle{IEEEtran}
\bibliography{mybib_normal}

\end{document}